\documentclass[sigconf%,anonymous
]{aamas} 

\usepackage{balance} % for balancing columns on the final page

\usepackage{savesym}
\savesymbol{Bbbk}
\savesymbol{openbox}

\restoresymbol{NEW}{Bbbk}
\restoresymbol{NEW}{openbox}
\usepackage{tcolorbox}

\usepackage{booktabs}
\usepackage{todonotes}
\usepackage{amsthm}
\usepackage{algorithm}
\usepackage{algorithmic}
\usepackage{multirow}
\usepackage{multicol}
\usepackage{thmtools}
\doi{NWEK7590}

\setcopyright{ifaamas}
\acmConference[AAMAS '26]{Proc.\@ of the 25th International Conference
on Autonomous Agents and Multiagent Systems (AAMAS 2026)}{May 25 -- 29, 2026}
{Paphos, Cyprus}{C.~Amato, L.~Dennis, V.~Mascardi, J.~Thangarajah (eds.)}
\copyrightyear{2026}
\acmYear{2026}
\acmDOI{}
\acmPrice{}
\acmISBN{}

\title[ArgLLM-App]{ArgLLM-App: An Interactive System for \\ Argumentative Reasoning with Large Language Models}
\author{Adam Dejl$^*$}\thanks{$^*$These authors contributed equally.
We thank Pranay Padavala for his early input.}
\affiliation{
  \institution{Imperial College London}
  \city{London}
  \country{United Kingdom}}
\email{adam.dejl18@imperial.ac.uk}

\author{Deniz Gorur$^*$}
\affiliation{
  \institution{Imperial College London}
  \city{London}
  \country{United Kingdom}}
\email{d.gorur22@imperial.ac.uk}

\author{Francesca Toni}
\affiliation{
  \institution{Imperial College London}
  \city{London}
  \country{United Kingdom}}
\email{ft@imperial.ac.uk}
\begin{abstract}

Argumentative LLMs (ArgLLMs) are an existing approach leveraging Large Language Models (LLMs) and computational argumentation for decision-making, with the aim of making the resulting decisions faithfully explainable to and contestable by humans. Here we propose a web-based system implementing ArgLLM-empowered agents for binary tasks. ArgLLM-App supports visualisation of the produced explanations and interaction with human users, allowing them to identify and contest any mistakes in the system's reasoning. It is highly modular and enables drawing information from trusted external sources. ArgLLM-App is publicly available at \href{https://argllm.app}{argllm.app}, with a video demonstration at \href{https://youtu.be/vzwlGOr0sPM}{youtu.be/vzwlGOr0sPM}.
\end{abstract}


\newcommand{\BibTeX}{\rm B\kern-.05em{\sc i\kern-.025em b}\kern-.08em\TeX}

\begin{document}

%%% The following commands remove the headers in your paper. For final 
%%% papers, these will be inserted during the pagination process.

\pagestyle{fancy}
\fancyhead{}

%%% The next command prints the information defined in the preamble.

\maketitle 

%%%%%%%%%%%%%%%%%%%%%%%%%%%%%%%%%%%%%%%%%%%%%%%%%%%%%%%%%%%%%%%%%%%%%%%%

\section{Introduction}
\label{sec:introduction}

Large Language Models (LLMs) have recently emerged as a major class of AI models. Specifically, the diversity of knowledge encoded in LLMs and their ability to apply this knowledge zero-shot in a range of settings makes them promising candidates for use in decision-making. However, they are currently limited by their inability to reliably provide outputs that are faithfully explainable and contestable when mistaken. \citet{Freedman2025ArgLLMs} introduced \emph{Argumentative LLMs} (\emph{ArgLLMs} in short) to reconcile these strengths and weaknesses. ArgLLMs can utilise (arbitrary) LLMs to construct 
quantitative bipolar argumentation frameworks (QBAFs)~\cite{rago2016discontinuity}, which then serve as the basis for formal reasoning in decision-making supported by gradual semantics for QBAFs~\cite{BaroniRT18}. The interpretable nature of these argumentation frameworks and formal reasoning means that any decision made by ArgLLMs may be faithfully explained to, and contested by, humans. To date, ArgLLMs have been proven effective, as well as competitive with  state-of-the-art techniques such as chain-of-thought~\cite{cot22}, in the binary decision-making task of 
claim verification.  

Our main contribution in this demo paper is  a system implementing ArgLLM-empowered agents  for  any 
binary decision-making task (including claim verification as in \cite{Freedman2025ArgLLMs}). 
We call the system \emph{ArgLLM-App} to emphasise that it can be naturally used to build applications with ArgLLMs.
ArgLLM-App is a web application supported by a server implementing the ArgLLM logic and mediating access to the base LLM.
It is highly flexible with respect to the depth and breadth of generated QBAFs, and the gradual semantics for the evaluation of opinions in these QBAFs.
The system also supports interaction with human users, enabling them to
modify aspects of the generated QBAF, either by adjusting the base confidence in arguments (in QBAF terminology: the arguments' base scores) or expanding the QBAF (in QBAF terminology: adding attackers/supporters of arguments).
This can be done directly via a QBAF visualisation
or indirectly through a chat interface.
Such modifications
lead ArgLLM-App to revise its assessment of the binary decision-making task at hand.
Human users
can also upload trusted sources, inputted as PDFs, to inform the generation of QBAFs, 
in the spirit of Retrieval-Augmented Generation (RAG)~\cite{rag20}.

\begin{figure*}[ht!]
    \centering
    \includegraphics[width=\linewidth]{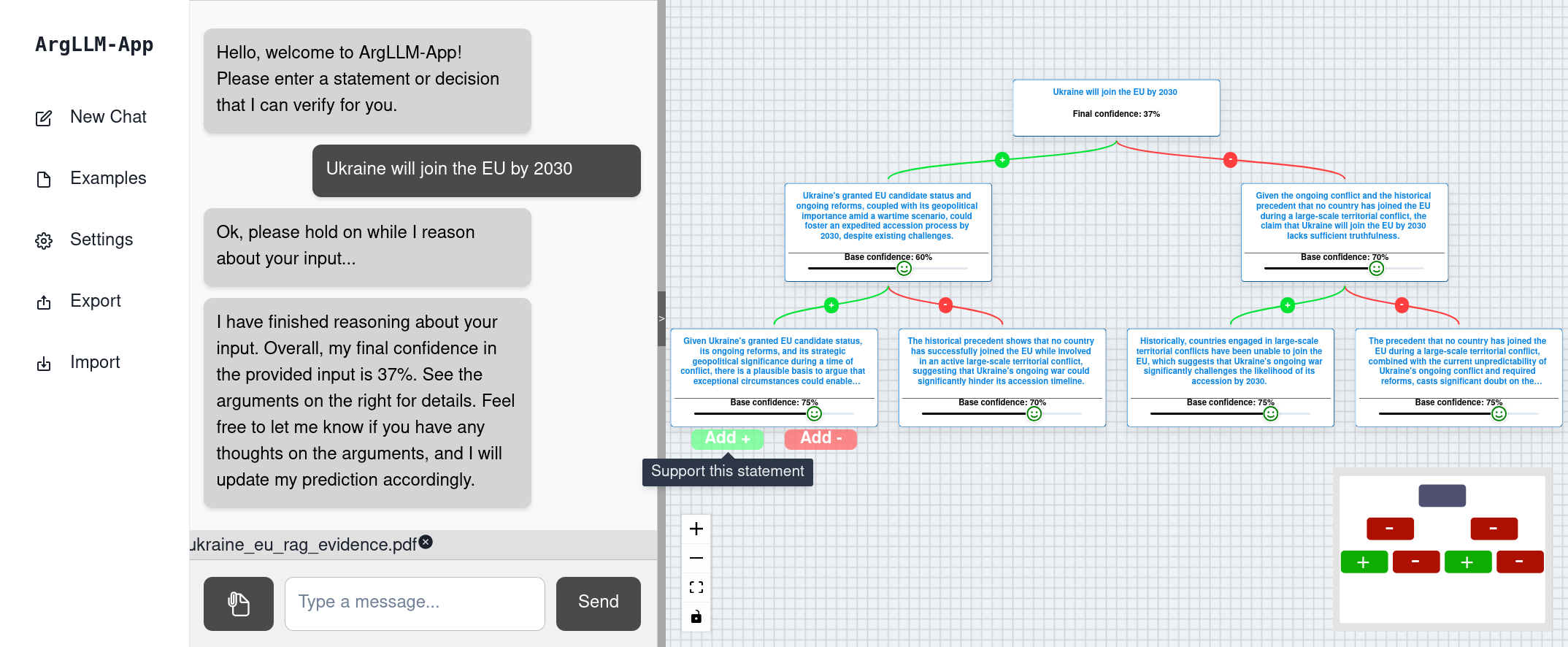}
    \caption{Output of ArgLLM-App for claim 
    ``Ukraine will join the EU before 2030'',
    with settings as in Figure~\ref{fig:settings}
    and document-based QBAF generation (see the PDF document indication at the bottom of the chat). In the QBAF, the bottom left argument has been selected
    for addition of a supporter at depth 3. The abstract graph at the bottom right shows the  QBAF structure. }
    \label{fig:argllm_d2_rag}
\end{figure*}

%%%%%%%%%%%%%%%%%%%%%%%%%%%%%%%%%%%%%%%%%%%%%%%%%%%%%%%%%%%%%%%%%%%%%%%%

\section{Preliminaries}
\label{sec:preliminaries}

\emph{QBAFs}~\cite{BaroniRT18}
\label{def:qbaf}
are quadruples $
(\mathcal{A}, \mathcal{R}^{-}, \mathcal{R}^{+}, \tau
)$ consisting of a finite set of \emph{arguments} $\mathcal{A}$, disjoint binary relations of \emph{attack} $\mathcal{R}^- \!\subseteq\! \mathcal{A} \!\times \!\mathcal{A}$ and \emph{support} $\mathcal{R}^+ \!\subseteq \!\mathcal{A} \!\times\! \mathcal{A}$, 
and a \emph{base score function} $\tau:\mathcal{A}\! \rightarrow \! [0,1]$.

QBAFs can be seen as graphs with arguments 
as nodes and elements of the attack and support relations as edges.
ArgLLMs adopt a graphical visualisation as illustrated in
Figure~\ref{fig:argllm_d2_rag}, right. Here, the QBAF has seven arguments (given by boxes) and attack and support relations (given by red/- and green/+ edges, respectively)
with three elements each.
The arguments are a claim (``Ukraine will join the EU before 2030''), evidence for/against the claim, or evidence for/against other evidence. 
The base scores are indicated as the \emph{base confidence}: their value is determined by the underpinning LLM by direct questioning~\cite{Freedman2025ArgLLMs}.
In the figure, as in ArgLLMs, the \emph{final confidence} of the claim argument is computed by a \emph{gradual semantics}. For QBAFs, this is a function
$\sigma:\! \mathcal{A} \!\rightarrow \! [0,1]$, often involving an iterative procedure initialising strength values with the base scores and 
then repeatedly 
updating
the strength of arguments based on the strength
of their attackers and supporters.
In the figure, $\sigma$ is DF-QuAD~\cite{rago2016discontinuity}, but ArgLLM-App 
also accommodates other semantics.

In ArgLLMs,
QBAFs, seen graphically, are restricted to trees with claims as their root (as in Figure~\ref{fig:argllm_d2_rag}). When they are generated with the help of LLMs, they are either of \emph{depth 1} (with a single layer of attackers and supporters for the claim) or of \emph{depth 2} (with each attacker or supporter in turn having its own attacker and supporter, as in Figure~\ref{fig:argllm_d2_rag}). 
In both cases,
in ArgLLMs, arguments have a single supporter and a single attacker (\emph{breadth 1}).
ArgLLM-App allows a choice between depth 1 and 2 options as well as a choice of 
breadth. Note that we limit the depth to 2 to avoid overloading the users.

%%%%%%%%%%%%%%%%%%%%%%%%%%%%%%%%%%%%%%%%%%%%%%%%%%%%%%%%%%%%%%%%%%%%%%%%

% \section{Related Work}
% \label{sec:related}

%%%%%%%%%%%%%%%%%%%%%%%%%%%%%%%%%%%%%%%%%%%%%%%%%%%%%%%%%%%%%%%%%%%%%%%%

\section{ArgLLM-App Features}
\label{sec:system}

ArgLLM-App is highly customisable. It allows the choice of the following
parameters:

\begin{itemize}
    \item QBAF depth (1 or 2);
    \item QBAF breadth (up to 4, i.e. up to 4 attackers and 4 supporters per argument, the same number);
    \item gradual semantics (DF-QuAD~\cite{rago2016discontinuity}, Euler-based~\cite{Amgoud17Euler} and Quadratic Energy~\cite{Potyka_18}).
\end{itemize}

\begin{figure}[tp!]
    \centering
    \includegraphics[width=0.48\linewidth
    % 5.3cm
    ]{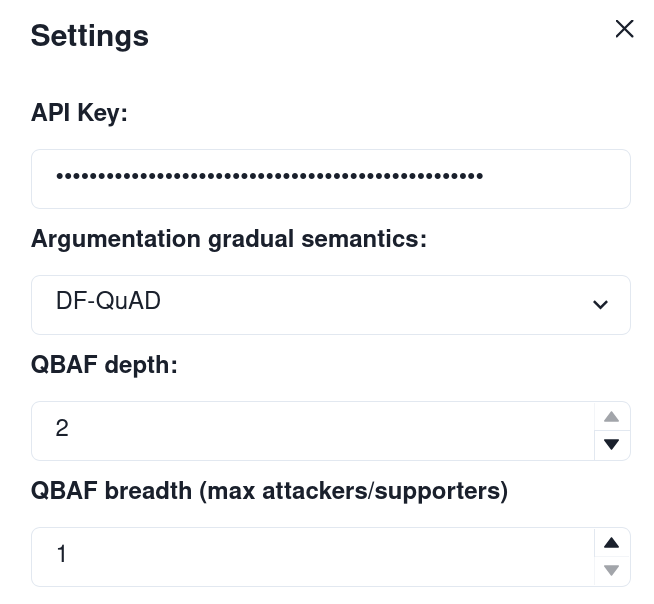}
    \caption{Settings: in addition to the API key for access to the base LLM, the user can configure the gradual semantics, depth and breadth to be used by ArgLLM-App.}
    \label{fig:settings}
\end{figure}

The values of the parameters can be configured before interacting with the system (see Figure~\ref{fig:settings}).
For binary decisions given in the chat, ArgLLM-App generates outputs as illustrated in Figure~\ref{fig:argllm_d2_rag}.

ArgLLM-App is highly interactive:
\begin{itemize}
\item 
Users can modify the generated outputs by using the base confidence slider to express what they deem to be the correct confidence in evidence for/against claims or (in the case of depth 2) other evidence.
\item 
Users can add additional supporters or attackers by using the Add buttons associated with arguments (Figure~\ref{fig:argllm_d2_rag} illustrates how to initiate the addition of a supporter).
    \item Users can
    augment the internal LLM knowledge through
    document-based QBAF generation with documents specified in a PDF format. These PDFs are parsed to MarkDown format and incorporated into the LLM prompts.
    Alternatively, users can provide
    additional information in the chat, which 
    ArgLLM-App can autonomously turn into attackers or supporters of the relevant arguments.
\end{itemize}

%%%%%%%%%%%%%%%%%%%%%%%%%%%%%%%%%%%%%%%%%%%%%%%%%%%%%%%%%%%%%%%%%%%%%%%%

\section{Future Work}
\label{sec:conclusions}
ArgLLM-App is a prototype that can be extended in several directions.
The system could be adjusted to support QBAF depths higher than 2 while avoiding cognitive overload of users.
It could also be enhanced to enable the upload of documents in formats other than PDF.
Other ways to compute base  confidence may also be useful~\cite{Kevin}.
The current realisation allows the use of only base LLMs from OpenAI, and only single LLMs.
We envisage the support of LLMs from other providers, and multi-agent variants where different agents rely upon different LLMs, in the spirit of 
\cite{deniz-aamas}.
We are currently allowing debate of a 
single binary decision: multiple decisions/question answering would also be useful. 
Further, it would be interesting to fully integrate RAG~\cite{rag20} into ArgLLM-App, 
allowing our agents to find relevant sources autonomously
and extract relevant arguments therefrom, as in \cite{deniz-aamas}. 
Finally, we accommodate interactions with single users: multiple users interacting concurrently with the system and each other may bring additional value.

%%%%%%%%%%%%%%%%%%%%%%%%%%%%%%%%%%%%%%%%%%%%%%%%%%%%%%%%%%%%%%%%%%%%%%%%

%%% The next two lines define, first, the bibliography style to be 
%%% applied, and, second, the bibliography file to be used.

\clearpage \newpage 

\section*{Acknowledgements}
This research was partially supported by ERC under the EU's Horizon 2020 research and innovation programme (grant agreement No. 101020934, ADIX). We thank all members of the Computational Logic and Argumentation group at Imperial College London, and in particular Antonio Rago and Gabriel Freedman, for their suggestions and feedback on earlier versions of this work. We also thank Pranay Padavala for his early contributions. 

\bibliographystyle{ACM-Reference-Format} 
\bibliography{bibliography}

@inproceedings{rago2016discontinuity,
  author       = {Antonio Rago and
                  Francesca Toni and
                  Marco Aurisicchio and
                  Pietro Baroni},
  opteditor       = {Chitta Baral and
                  James P. Delgrande and
                  Frank Wolter},
  title        = {Discontinuity-Free Decision Support with Quantitative Argumentation
                  Debates},
  optbooktitle    = {Principles of Knowledge Representation and Reasoning: Proceedings
                  of the Fifteenth International Conference, {KR} 2016, Cape Town, South
                  Africa, April 25-29, 2016},
  booktitle    = {{KR} 2016},
  optpages        = {63--73},
  publisher    = {{AAAI} Press},
  year         = {2016},
  url          = {http://www.aaai.org/ocs/index.php/KR/KR16/paper/view/12874}
}

@inproceedings{Amgoud17Euler,
  author       = {Leila Amgoud and
                  Jonathan Ben{-}Naim},
  opteditor       = {Alessandro Antonucci and
                  Laurence Cholvy and
                  Odile Papini},
  title        = {Evaluation of Arguments in Weighted Bipolar Graphs},
  optbooktitle    = {Symbolic and Quantitative Approaches to Reasoning with Uncertainty
                  - 14th European Conference, {ECSQARU} 2017, Lugano, Switzerland, July
                  10-14, 2017, Proceedings},
  booktitle    = {{ECSQARU} 2017},
  optseries       = {Lecture Notes in Computer Science},
  optvolume       = {10369},
  optpages        = {25--35},
  publisher    = {Springer},
  year         = {2017},
  url          = {https://doi.org/10.1007/978-3-319-61581-3\_3},
  doi          = {10.1007/978-3-319-61581-3\_3}
}

@inproceedings{BaroniRT18,
  author       = {Pietro Baroni and
                  Antonio Rago and
                  Francesca Toni},
  title        = {How Many Properties Do We Need for Gradual Argumentation?},
  booktitle    = {{AAAI} 2018},
  optpages        = {1736--1743},
  publisher    = {{AAAI} Press},
  year         = {2018},
  doi          = {10.1609/aaai.v32i1.11544}
}

@inproceedings{Freedman2025ArgLLMs,
  author       = {Gabriel Freedman and
                  Adam Dejl and
                  Deniz Gorur and
                  Xiang Yin and
                  Antonio Rago and
                  Francesca Toni},
  opteditor       = {Toby Walsh and
                  Julie Shah and
                  Zico Kolter},
  title        = {Argumentative Large Language Models for Explainable and Contestable
                  Claim Verification},
  optbooktitle    = {AAAI-25, February 25 - March 4, 2025, Philadelphia, PA, {USA}},
  booktitle     = {{AAAI} 2025},
  optpages        = {14930--14939},
  publisher    = {{AAAI} Press},
  year         = {2025},
  url          = {https://doi.org/10.1609/aaai.v39i14.33637},
  doi          = {10.1609/AAAI.V39I14.33637}
}

@inproceedings{cot22,
  author       = {Jason Wei and
                  Xuezhi Wang and
                  Dale Schuurmans and
                  Maarten Bosma and
                  Brian Ichter and
                  Fei Xia and
                  Ed H. Chi and
                  Quoc V. Le and
                  Denny Zhou},
  title        = {Chain-of-Thought Prompting Elicits Reasoning in Large Language Models},
  booktitle    = {NeurIPS 2022},
  year         = {2022},
  url          = {http://papers.nips.cc/paper\_files/paper/2022/hash/9d5609613524ecf4f15af0f7b31abca4-Abstract-Conference.html}
}

@inproceedings{rag20,
  author       = {Patrick Lewis and
                  Ethan Perez and
                  Aleksandra Piktus and
                  Fabio Petroni and
                  Vladimir Karpukhin and
                  Naman Goyal and
                  Heinrich K{\"{u}}ttler and
                  Mike Lewis and
                  Wen{-}tau Yih and
                  Tim Rockt{\"{a}}schel and
                  Sebastian Riedel and
                  Douwe Kiela},
  title        = {Retrieval-Augmented Generation for Knowledge-Intensive {NLP} Tasks},
  booktitle    = {NeurIPS 2020},
  year         = {2020},
  url          = {https://proceedings.neurips.cc/paper/2020/hash/6b493230205f780e1bc26945df7481e5-Abstract.html}
}

@inproceedings{Potyka_18,
  author       = {Nico Potyka},
  opteditor       = {Michael Thielscher and
                  Francesca Toni and
                  Frank Wolter},
  title        = {Continuous Dynamical Systems for Weighted Bipolar Argumentation},
  booktitle    = {KR 2018},
  optpages        = {148--157},
  publisher    = {{AAAI} Press},
  year         = {2018},
  url          = {https://aaai.org/ocs/index.php/KR/KR18/paper/view/17985}
}

@inproceedings{deniz-aamas,
  author       = {Deniz Gorur and Antonio Rago and Francesca Toni},
      title        = {Retrieval- and Argumentation-Enhanced Multi-Agent LLMs for
Judgmental Forecasting},
      optbooktitle    = {{AAMAS} 2026 - 25th International Conference on Autonomous Agents and Multiagent Systems, 25-29 May 2026, Paphos, Cyprus},
      booktitle    = {AAMAS 2026},
      year         = {2026},
  url          = {https://doi.org/10.65109/SNBR1486},
  doi          = {10.65109/SNBR1486},
  publisher    = {{International Foundation for Autonomous Agents and Multiagent Systems
                  / {ACM}}}
}

@inproceedings{kevin,
  author       = {Kevin Zhou and
                  Adam Dejl and
                  Gabriel Freedman and
                  Lihu Chen and
                  Antonio Rago and
                  Francesca Toni},
  title        = {Evaluating Uncertainty Quantification Methods in Argumentative Large
                  Language Models},
  publisher = {Association for Computational Linguistics},
  booktitle      = {EMNLP 2025},
  year         = {2025},
  doi = "10.18653/v1/2025.findings-emnlp.1184",
}

%%%%%%%%%%%%%%%%%%%%%%%%%%%%%%%%%%%%%%%%%%%%%%%%%%%%%%%%%%%%%%%%%%%%%%%%

\iffalse
\clearpage
\newpage
\section*{Demo Requirements}

To present the demo, we will need a desk, a power outlet (ideally with an extension cord) and a screen with an HDMI cable.
%We will need a desk and a screen with a HDMI cable.

\clearpage
\newpage
\onecolumn

\section*{Supervisor Endorsement}

The first two authors are PhD students. The last author is their supervisor.
The supervisor declares that the system demoed was fully developed by the PhD students in discussion with the supervisor herself.

Signed: 
Francesca Toni

\fi

\end{document}